\documentclass{article} % For LaTeX2e
\usepackage{iclr2023_conference,times}
\usepackage{graphicx}

\usepackage{amsmath,amsfonts,bm}

\def\eqref#1{equation~\ref{#1}}
\def\1{\bm{1}}

\DeclareMathAlphabet{\mathsfit}{\encodingdefault}{\sfdefault}{m}{sl}
\SetMathAlphabet{\mathsfit}{bold}{\encodingdefault}{\sfdefault}{bx}{n}

\usepackage{hyperref}
\usepackage{url}
\usepackage{subcaption}
\usepackage{wrapfig}
\usepackage{booktabs}
\usepackage{float}

\iclrfinalcopy

\title{GSL-PCD: Improving Generalist-Specialist Learning with Point Cloud Feature-based Task Partitioning}

\author{Xiu Yuan \\
University of California, San Diego \\
\texttt{x1yuan@ucsd.edu}}

\begin{document}

\maketitle

\begin{abstract}
Generalization in Deep Reinforcement Learning across unseen environment variations often requires training over a diverse set of scenarios. Many existing DRL algorithms struggle with efficiency when handling numerous environment variations. The Generalist-Specialist Learning (GSL) framework addresses this by first training a generalist on all variations, then creating specialists from the generalist’s weights, each focusing on a random subset of variations. Finally, the generalist refines its learning with assistance from the specialists. However, random task partitioning in GSL can impede specialist performance, as it often assigns vastly different variations to the same specialist, typically resulting in each specialist being assigned just one variation, which increases computational costs. To improve this, we propose Generalist-Specialist Learning with Point Cloud Feature-based Task Partitioning (GSL-PCD). This approach clusters environment variations based on features extracted from object point clouds, using balanced clustering with a greedy algorithm to assign similar variations to the same specialist. Evaluations on robotic manipulation tasks from the ManiSkill benchmark demonstrate that point cloud feature-based partitioning outperforms vanilla partitioning by 9.4\% with a fixed number of specialists and reduces computational and sample requirements by 50\% to achieve comparable performance.
\end{abstract}

\section{Introduction}
Generalization remains one of the most fundamental challenges in reinforcement learning (RL) \cite{cobbe2020leveraging, Jia2022ImprovingPO, ghosh2021generalization}. Ideally, RL agents should be capable of adapting and generalizing to unseen environment variations during deployment. To achieve this objective, RL agents need to focus on learning generalizable skills, rather than overfitting to specific environment setups. Recently, several benchmarks \cite{james2020rlbench, ghosh2021generalization, Gu2023ManiSkill2AU} have been proposed to evaluate the generalization capabilities of RL agents, offering rich scene layouts, a large number of environment configurations, and diverse sets of objects.

Many existing deep reinforcement learning (DRL) algorithms struggle to achieve high performance when faced with a wide range of environment variations, as demonstrated in \cite{Schulman2017ProximalPO}. To address these challenges, the Generalist-Specialist Learning (GSL) framework \cite{Jia2022ImprovingPO} was recently proposed. Inspired by how human organizations tackle complex problems, GSL first trains a \textit{generalist} on all environment variations. It then deploys a large population of \textit{specialists}, initialized with the generalist's weights, each trained on a randomly selected subset of variations. Finally, GSL retrains the \textit{generalist}, leveraging demonstrations from all the specialists.

During specialist training, GSL randomly assigns environment variations to specialists (\textbf{random task partitioning}). While this approach is convenient to implement, our empirical observations show that it often assigns very different environment variations to a single specialist, significantly increasing the difficulty of specialist training. As a result, to ensure specialists can effectively master their assigned variations, the number of environment variations per specialist should be reduced. In practice, GSL assigns only one environment variation per specialist in most of its experiments. However, this low number of variations per specialist leads to an explosion in the number of specialists required, resulting in an unsustainable demand for online samples and computational resources. For instance, GSL trains 75 specialists for Procgen tasks \cite{cobbe2020leveraging} and requires 1,024 specialists for the Plunder environment to achieve good performance. This level of sample and computational complexity severely limits GSL’s applicability in scenarios where online samples are costly and computational power is constrained.

We draw inspiration from how humans learn to solve complex tasks. People typically begin by practicing on a group of similar task configurations, rather than randomly selected ones, and only after becoming proficient in that group do they move on to broader sets of tasks. Based on this observation, we believe the principle of "starting with similar tasks first" can also be applied to the domain of Generalist-Specialist Learning (GSL). During specialist training, specialists should focus on subsets of similar task configurations to achieve better performance. In robotic manipulation tasks involving object variations, object appearance serves as the best indicator of task similarity \cite{ling2023efficacy}. Previous works like \cite{ling2023efficacy, he2021ffb6d, eisner2022flowbot3d} have noted that point cloud features encode detailed depth and spatial information, providing the most effective 3D representations for reasoning about agent-object relationships. Therefore, we naturally ask: \textbf{Can object point cloud features be leveraged to explore better task partitioning in Generalist-Specialist Learning (GSL) in robotic manipulation tasks with object variations?}

The answer is yes. In this paper, we propose Generalist-Specialist Learning with Point Cloud Feature-based Task Partitioning (GSL-PCD). Our method leverages a pre-trained PointNet++ model \cite{Qi2017PointNetDH} to encode point cloud features of all objects and employs a balanced clustering algorithm with a greedy approach to group objects with similar point cloud representations, assigning each cluster to a single specialist. By using point cloud features as an indicator of task similarity, our approach ensures that objects with similar characteristics are grouped under the same specialist. This significantly reduces the learning difficulty for each specialist, enabling them to handle a greater variety of environment variations.

We evaluate our method on the Turn Faucet task from the ManiSkill benchmark \cite{Gu2023ManiSkill2AU, mu2021maniskill}, a challenging robotic manipulation task featuring 60 different types of faucets. Empirical results demonstrate that our approach outperforms random partitioning by 9.4\% with a fixed number of specialists and reduces
computation and sample requirements by 50\% to achieve similar performance.

To summarize, our contributions are as follows:

\begin{itemize}
    \item We identify the limitations of random task partitioning within the vanilla Generalist-Specialist Learning (GSL) framework and advocate for a more structured approach to task partitioning.
    \item We propose Generalist-Specialist Learning with Point Cloud Feature-based Task Partitioning (GSL-PCD), which leverages point cloud features extracted by a pre-trained PointNet++ model to indicate task similarity in robotic manipulation tasks with object variations. Balanced clustering with greedy algorithm is used to group objects, assigning those from the same cluster to a single specialist.
    \item We evaluate our method on the challenging Turn Faucet task from the ManiSkill benchmark, featuring 60 different types of faucets, outperforms random partitioning by 9.4\% with a fixed number of specialists and reduces
computation and sample requirements by 50\% to achieve similar performance.
\end{itemize}

\section{Related Works}

\paragraph{\textbf{Divide-and-Conquer in RL}} The concept of employing a divide-and-conquer approach in training RL agents has been explored in previous works \cite{ghosh2017divide, teh2017distral}. These studies typically divide the state space into subsets, alternating between training specialist policies on each subset and using behavior cloning to transfer knowledge from the specialists to a generalist policy. Recently, Generalist-Specialist Learning (GSL) \cite{Jia2022ImprovingPO} systematically examined the timing and interaction between generalist and specialist training. GSL begins by training a generalist across all environment variations, then introduces a large group of specialists, each trained on a subset of variations, before resuming generalist training with assistance from the specialists' demonstrations. Our work extends GSL by enhancing the task partitioning process.

\paragraph{\textbf{Point Clouds in RL}} Numerous prior works have shown that point clouds are powerful visual representations in areas such as perception \cite{he2021ffb6d, vu2022softgroup}, self-driving \cite{wang2019pseudo, you2022learning}, and robotic manipulation \cite{eisner2022flowbot3d, simeonov2022neural}. Recently, reinforcement learning from 3D point clouds has been extensively studied in the literature \cite{ling2023efficacy, liu2022frame}. While our work does not directly incorporate point clouds into the RL training process, we leverage object point cloud features as indicators of task similarity in robotic manipulation tasks involving object variations.

\section{Background}

\begin{figure}[h]
    \centering
    \includegraphics[width=0.75\textwidth]{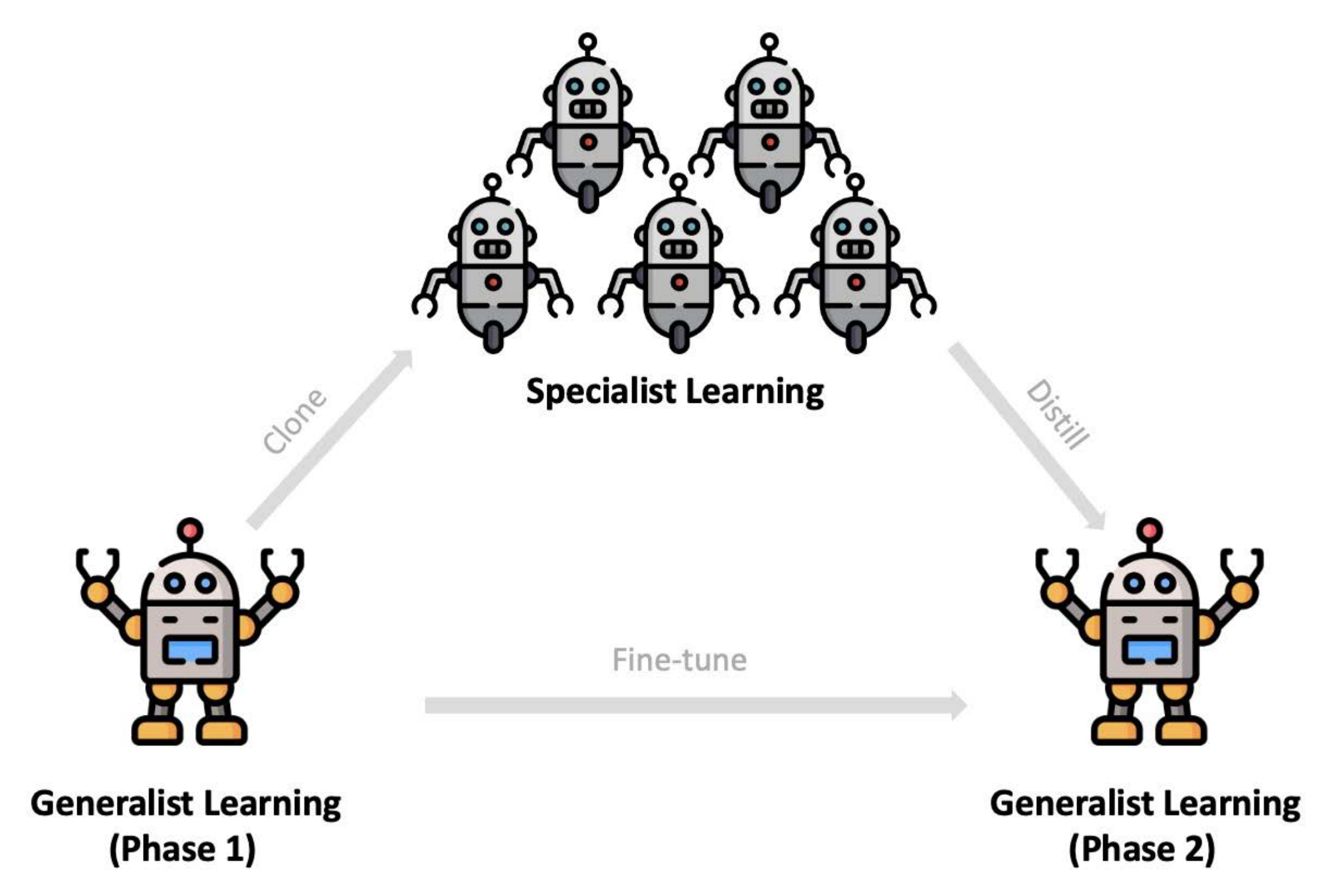}  % Replace with your image file
    \caption{Generalist-Specialist Learning framework}
    \label{fig:gsl}
\end{figure}

\paragraph{\textbf{Generalist-Specialist Learning}} As illustrated by \ref{fig:gsl}, Generalist-Specialist Learning (GSL) \cite{Jia2022ImprovingPO} follows the following steps:

\begin{enumerate}
    \item \textbf{Generalist Training}  The process begins by initializing a generalist policy, $\pi^g$, which is trained across all environment variations using RL algorithms such as PPO \cite{Schulman2017ProximalPO} and SAC \cite{Haarnoja2018SoftAO}. Once the performance reaches a plateau, the training of the generalist is paused, and the next phase is initiated.
    \item \textbf{Specialist Training}  Next, the $N_{low}$ lowest-performing environment variations are selected and randomly assigned to $N_s$ specialists, with their weights initialized by cloning from the generalist. The specialists are then trained in parallel, each for a fixed number of samples, $N_{sample}$.
    \item \textbf{Generalist Fine-tuning Guided by Specialists Demonstrations} Demonstrations are collected from the specialists on their assigned environment variations and from the generalist on the remaining variations. The generalist training is then resumed, utilizing learning-from-demonstration algorithms (such as DAPG+PPO \cite{Rajeswaran2017LearningCD} or GAIL+SAC \cite{Ho2016GenerativeAI}) guided by the collected demonstration set.
    
\end{enumerate}

\section{Problem Setup}

In this paper, we focus on enhancing Generalist-Specialist Learning (GSL) \cite{Jia2022ImprovingPO} through point cloud feature-based task partitioning. Our approach is grounded in two fundamental assumptions:

\begin{enumerate} 
    \item The environment involves object variations, such as turn faucet task with different types of faucets. 
    \item Point clouds for all objects are available.
\end{enumerate}

\section{Methods}

\begin{figure}[h]
    \centering
    \includegraphics[width=0.75\textwidth]{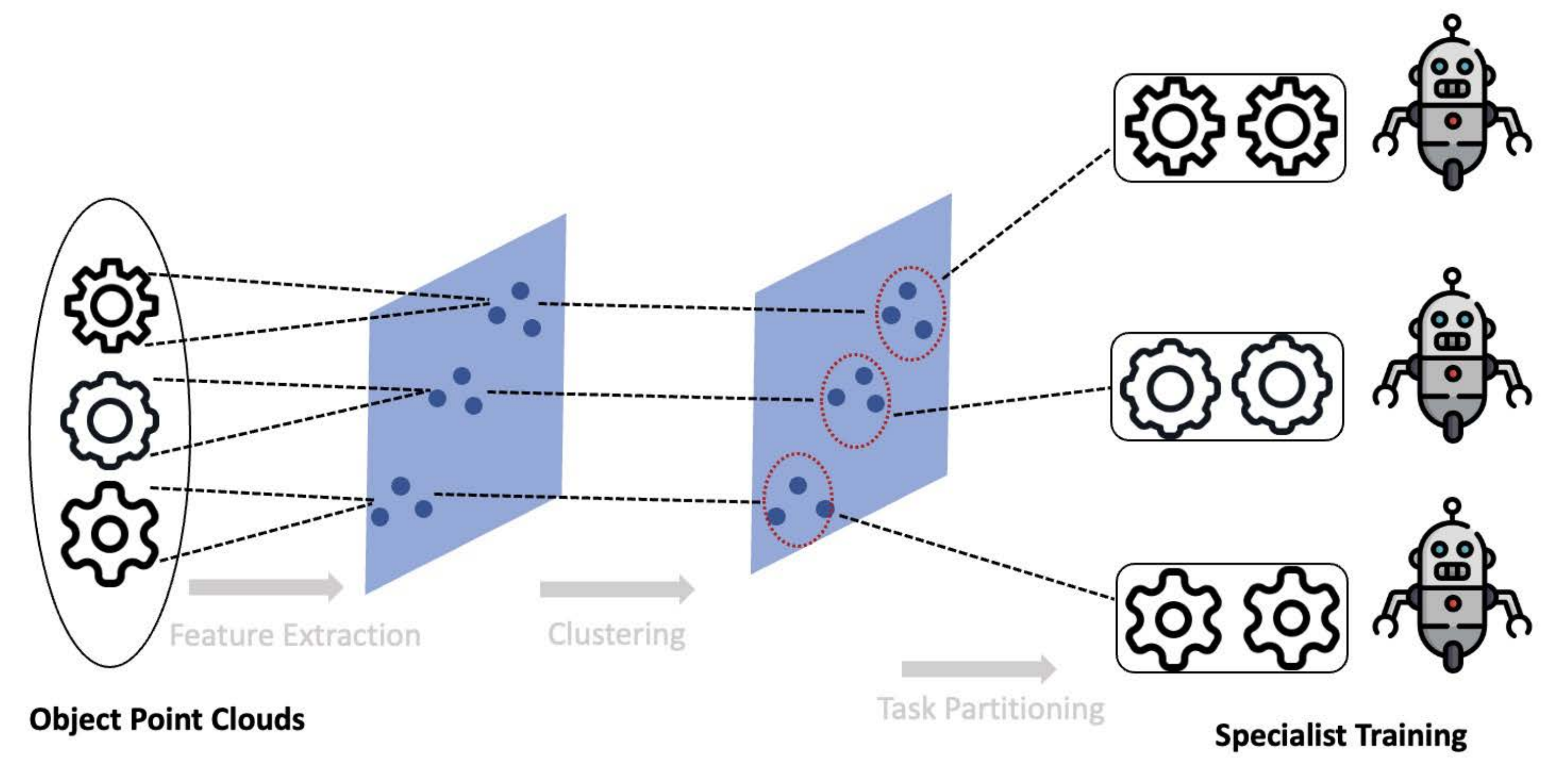}  % Replace with your image file
    \caption{Task Partitioning with Point Cloud Feature}
    \label{fig:gsl_pcd}
\end{figure}

As illustrated by \ref{fig:gsl_pcd}, we propose an improved task partitioning strategy that leverages point cloud features to enhance the efficiency of specialist training within the framework of Generalist-Specialist Learning (GSL), specially in environments with object variations. In Section \ref{sec:point_cloud_feature}, we discuss the method for extracting point cloud features. Section \ref{sec:clustering} covers the techniques for balanced clustering with greedy algorithm.

\subsection{Point Cloud Feature Extraction}
\label{sec:point_cloud_feature}

We assume that point clouds for all objects are available and utilize a pre-trained PointNet++ model \cite{Qi2017PointNetDH}, trained on the ShapeNet dataset \cite{chang2015shapenet}, for feature extraction with a feature vector dimension of 1024. We observe that many of the feature vector dimensions are zero, and the high dimensionality poses challenges for clustering algorithms. To address this, we normalize the features using Euclidean distance and apply Principal Component Analysis (PCA) \cite{abdi2010principal}, reducing the dimensionality to 2 and improving clustering performance. This pre-trained model enables efficient extraction of meaningful features from the object point clouds, which are then used to guide our task partitioning strategy.

\subsection{Balanced Clustering with Greedy Algorithm}
\label{sec:clustering}

After extracting features from the point clouds, we aim to partition environment variations using the KMeans algorithm \cite{ahmed2020k}, guided by the point cloud features. However, the standard KMeans approach does not ensure balanced partitioning, often resulting in clusters with uneven numbers of elements. Specifically, some clusters may contain significantly more environment variations than others, posing challenges for the corresponding specialists. To address this, we propose a balanced clustering method using a greedy algorithm. First, we compute $N_{cluster}$ centroids using the KMeans algorithm and maintain a distance table that records the distances between each pair of feature vectors and centroids. We then process these pairs in increasing order of distance. If the cluster associated with the centroid is not yet full, the feature vector is assigned to that cluster; otherwise, we move to the next pair. By applying this balanced clustering with a greedy approach, we partition the feature vectors into $N_{cluster}$ distinct groups. We then initialize $N_{cluster}$ specialists, with each specialist assigned to one of the clustered groups. This method ensures that each specialist focuses on a specific subset of tasks tailored to the characteristics of the objects in their group, while preventing any specialist from being overloaded with too many environment variations. See Section \ref{sec:balanced_clustering} for comparision results between vanilla clustering and our proposed balanced clustering with greedy algorithm.

\section{Experiments}

The goal of our experiment evaluation is to study the following questions:

\begin{enumerate} 
    \item Can point cloud feature-based task partitioning effectively reduce specialist training costs compared to vanilla GSL with random task partitioning? (\ref{sec:main_result})
    \item What are the effects of balanced clustering compared with vanilla clustering?(\ref{sec:balanced_clustering})
    \item Does point cloud feature-based task partitioning effectively group environment variations with high task similarity together, and how does this approach benefit specialist training? (\ref{sec:discussion})
\end{enumerate}

\subsection{Experiment Setup}
\label{sec:experiment_setup}

We conduct experiments using the turn faucet task from the ManiSkill benchmark \cite{mu2021maniskill}, which includes 60 different types of faucets. Our experiments utilize low-dimensional state observations, such as robot proprioception (joint angles, joint velocities, end-effector pose, base pose, etc.) and goal information. For each faucet type, we randomly sample 10,000 points from its surface for feature extraction. The turn faucet task environment provides human-engineered dense rewards. We use PPO \cite{Schulman2017ProximalPO} as the backbone RL algorithm for both generalist and specialist training, and apply PPO+DAPG \cite{Rajeswaran2017LearningCD} for generalist fine-tuning.

\subsection{Main Result}
\label{sec:main_result}

\begin{wrapfigure}{r}{0.4\linewidth}
    \centering
    \vspace{-0.8cm}
    \includegraphics[width=\linewidth]{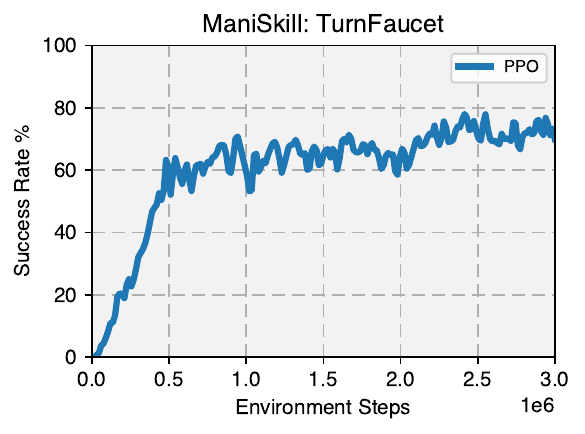}
    \vspace{-0.5cm}
    \caption{Generalist training (phase 1) using PPO on all environment variations (all types of faucets)}
    \label{fig:faucet_ppo}
    \vspace{-0.2cm}
\end{wrapfigure}

\begin{table}[h]
\centering
\begin{tabular}{lcccccccl}
\toprule
\textbf{Agent} & \textbf{Average} & \textbf{Median} & \textbf{High} & \textbf{Low} & \textbf{Upper Quantile} & \textbf{Lower Quantile} \\
\midrule
Generalist (Phase 1) & \textbf{73\%} & 91\% & 100\% & 0\% & 98\% & 53\% \\
\bottomrule
\end{tabular}
\caption{We evaluate the generalist (phase 1) on each of the 60 faucets over 100 episodes and report the success rate statistics}
\label{tab:generalist}
\end{table}

\subsubsection{Generalist Training (Phase 1)}
We first train the generalist on all variations of the environment (i.e., all types of faucets) using PPO, as illustrated in Figure \ref{fig:faucet_ppo}. After training the generalist for 3 million steps, its performance plateaus at approximately 70\%. Subsequently, we conduct full-scale evaluations of the generalist, as shown in Table \ref{tab:generalist}. The generalist is evaluated on each of the 60 faucets across 100 episodes, and we report the success rate statistics. 

The average success rate across all 60 faucets is 73\%, while the median success rate stands at 91\%. This indicates that the generalist model performs well on most faucets but struggles with a few. To address this, we focus on training the generalist on 29 faucets with performance rates below the median.

\subsubsection{Specialist Training}

\begin{table}[h]
\centering
\begin{tabular}{lcccl}
\toprule
\textbf{Agent} & \textbf{Number of Specialists} & \textbf{Average} \\
\midrule
Generalist (Phase 1) & --- & \textbf{46.5\%} \\
Specialists (Random Partitioning) & 4 & \textbf{75.6\%} \\
Specialists (Random Partitioning) & 6 & \textbf{80.1\%} \\
Specialists (Random Partitioning) & 8 & \textbf{83.4\%} \\
Specialists (Point Cloud Feature-based Partitioning) & 4 & \textbf{82.7\%} \\
\bottomrule
\end{tabular}
\caption{We evaluate the generalist (Phase 1) and all specialist groups under various task partitioning methods on 29 selected low-performing faucets over 100 episodes, reporting the average success rate}
\label{tab:main}
\end{table}

With 29 low-performance faucets selected, we explore two types of task partitioning—random and point cloud feature-based—for varying numbers of specialists. As demonstrated in Table \ref{tab:main}, the generalist in phase 1 achieves an average success rate of 46.5\% over selected 29 faucets. We train each specialist for 1 million steps with PPO and use the average success rate across 29 low-performing faucets as a key metric to evaluate the effectiveness of specialist training. Our experiments aim to address the following questions:

\begin{enumerate}
    \item \textbf{Does point cloud feature-based task partitioning achieve a higher average success rate compared to random task partitioning when the number of specialists is fixed (fixed number of specialists)?}
    \item \textbf{How many computes and samples can point cloud feature-based task partitioning save compared to random task partitioning to achieve a similar average success rate (fixed average success rate)?}
\end{enumerate}

\paragraph{\textbf{Fixed Number of Specialists}} We set the number of specialists to 4 and compare the average success rate achieved by random partitioning and point cloud feature-based partitioning. As shown in Table \ref{tab:main}, with 4 specialists, random partitioning yields an average success rate of 75.6\%, while our approach achieves an average success rate of 82.7\%. This represents a significant improvement of 9.4\% in average success rate over 29 selected faucets when compared to random partitioning.

\paragraph{\textbf{Fixed Average Sucess Rate}} We attempt to increase the number of specialists in random partitioning to determine how many are needed to achieve performance similar to that of point cloud feature-based partitioning. As shown in Table \ref{tab:main}, increasing the number of specialists to 8 allows random partitioning to reach an average success rate of 83.4\%, comparable to the 82.7\% achieved by point cloud feature-based partitioning. This experiment clearly demonstrates that point cloud feature-based partitioning reduces computational and sample requirements by 50\%.

\subsubsection{Generalist Fine-tuning (Phase 2)}

After completing specialist training, we collect 10 demonstrations for each faucet from both the generalist and the specialists. For the 29 low-performance faucets that underwent specialist training, demonstrations are gathered from their respective specialists. For the remaining high-performance faucets, demonstrations are collected from the generalist. In total, the demonstration dataset now comprises 600 trajectories. We then fine-tune the generalist using the demonstration dataset with PPO+DAPG \cite{Rajeswaran2017LearningCD}. This section aims to address the following questions through our experiments:

\begin{enumerate}
    \item \textbf{When the number of specialists is fixed, does a higher average success rate achieved through point cloud feature-based task partitioning lead to better final performance in generalist fine-tuning compared to random task partitioning?}
    \item \textbf{With similar average performance, does point cloud feature-based task partitioning achieve comparable final performance to random task partitioning which uses a greater number of specialists?}
\end{enumerate}

As shown in Table \ref{tab:main} and Fig, the average success rate across the 29 low-performing faucets during specialist training proves to be a reliable predictor of generalist fine-tuning performance in Phase 2. Point cloud feature-based partitioning with 4 specialists achieves similar performance to random partitioning with 8 specialists in the generalist fine-tuning phase and significantly outperforms random partitioning with 4 specialists. This observation aligns with our findings in Section \ref{sec:main_result}, further reinforcing the effectiveness and necessity of our approach.

\subsection{Effects of Balanced Clustering}
\label{sec:balanced_clustering}

\begin{table}[h]
\centering
\begin{tabular}{lcccl}
\toprule
\textbf{Agent} & \textbf{Environment Variations in Each Cluster} & \textbf{Average} \\
\midrule
Specialists (Vanilla Clustering) & (4, 6, 9, 10) & \textbf{75.8\%} \\
Specialists (Balanced Clustering) & (7, 7, 7, 8) & \textbf{82.7\%} \\
\bottomrule
\end{tabular}
\caption{We compare balanced clustering with vanilla clustering using 4 specialists on 29 selected low-performing faucets over 100 episodes, reporting the number of faucets in each cluster and the average success rate}
\label{tab:balanced_clustering}
\end{table}

\begin{table}[h]
\centering
\begin{tabular}{lcccl}
\toprule
\textbf{Agent} & \textbf{Number of Environment Variations} & \textbf{Average} \\
\midrule
Specialist 1 & 4 & \textbf{86\%} \\
Specialist 2 & 6 & \textbf{76\%} \\
Specialist 3 & 9 & \textbf{74\%} \\
Specialist 4 & 10 & \textbf{68\%} \\
\bottomrule
\end{tabular}
\caption{The performance discrepancy of 4 specialists of \textbf{vanilla clustering} evaluated for 100 episodes on their corresponding faucets}
\label{tab:vanilla_partitioning}
\end{table}

In this section, we aim to evaluate the impact of balanced clustering using the greedy algorithm. As shown in Table \ref{tab:balanced_clustering}, vanilla clustering results in an uneven distribution of environment variations, with the 4 clusters containing 4, 6, 9, and 10 faucets, respectively. Consequently, vanilla clustering achieves an average success rate of only 75.8\%, which is comparable to the performance of random partitioning. In contrast, balanced clustering achieves an improved average success rate of 82.7\%. As illustrated in Table \ref{tab:vanilla_partitioning}, the unequal distribution of specialists in vanilla clustering leads to significant performance discrepancies across different specialists, with those handling more environment variations struggling, achieving only 68\%. This experiment clearly demonstrates the effectiveness and necessity of balanced partitioning for achieving optimal performance.

\subsection{Discussion on point cloud feature-based task partitioning}
\label{sec:discussion}

This section aims to determine whether point cloud feature-based task partitioning effectively groups similar object variations and how it enhances specialist learning.

\begin{figure}[h]
    \centering
    \includegraphics[width=0.95\textwidth]{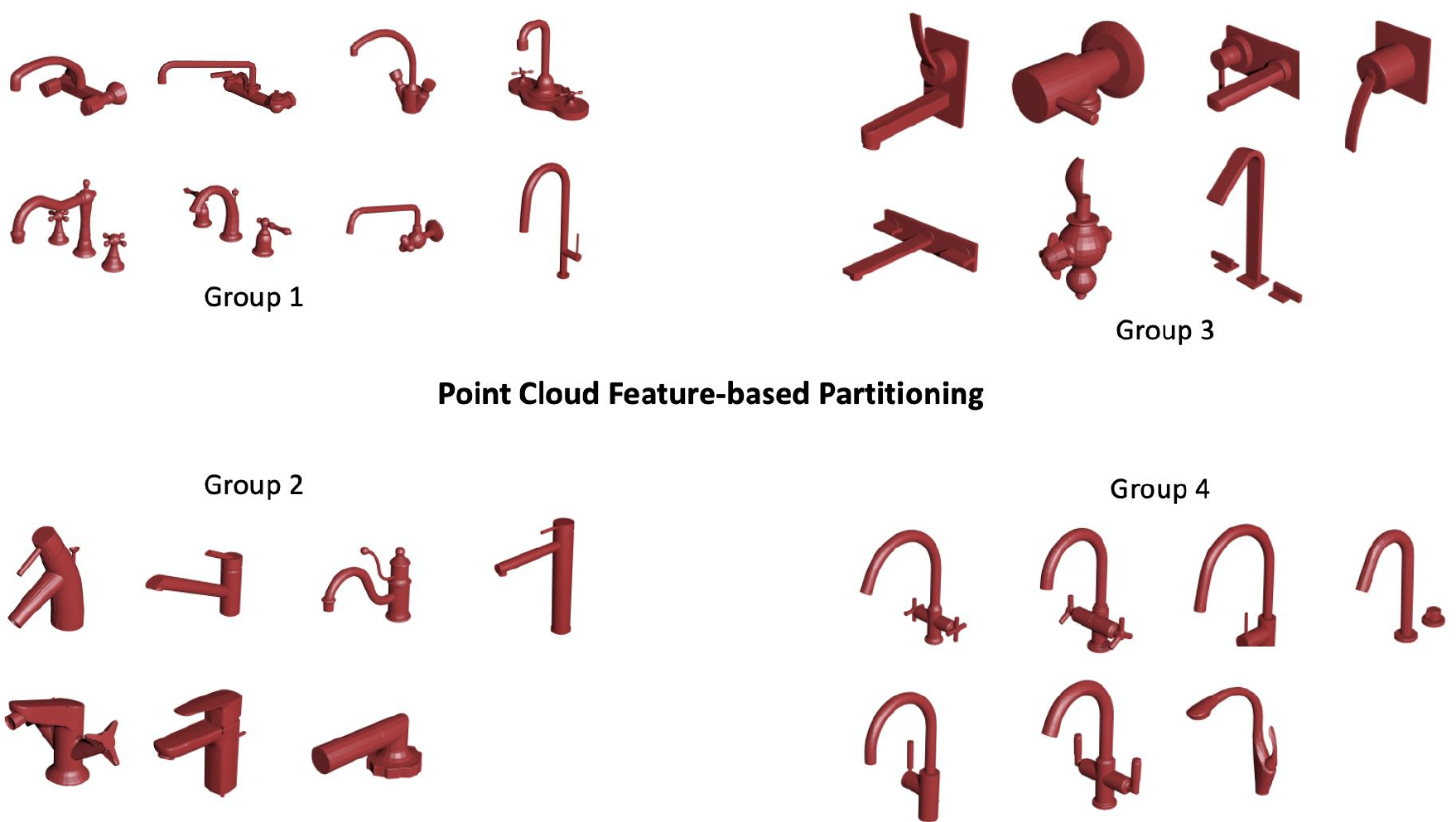}  % Replace with your image file
    \caption{An illustration of how point cloud feature-based partitioning separates 29 low-performing faucets into 4 groups}
    \label{fig:point_cloud_partition}
\end{figure}

\begin{figure}[h]
    \centering
    \includegraphics[width=0.95\textwidth]{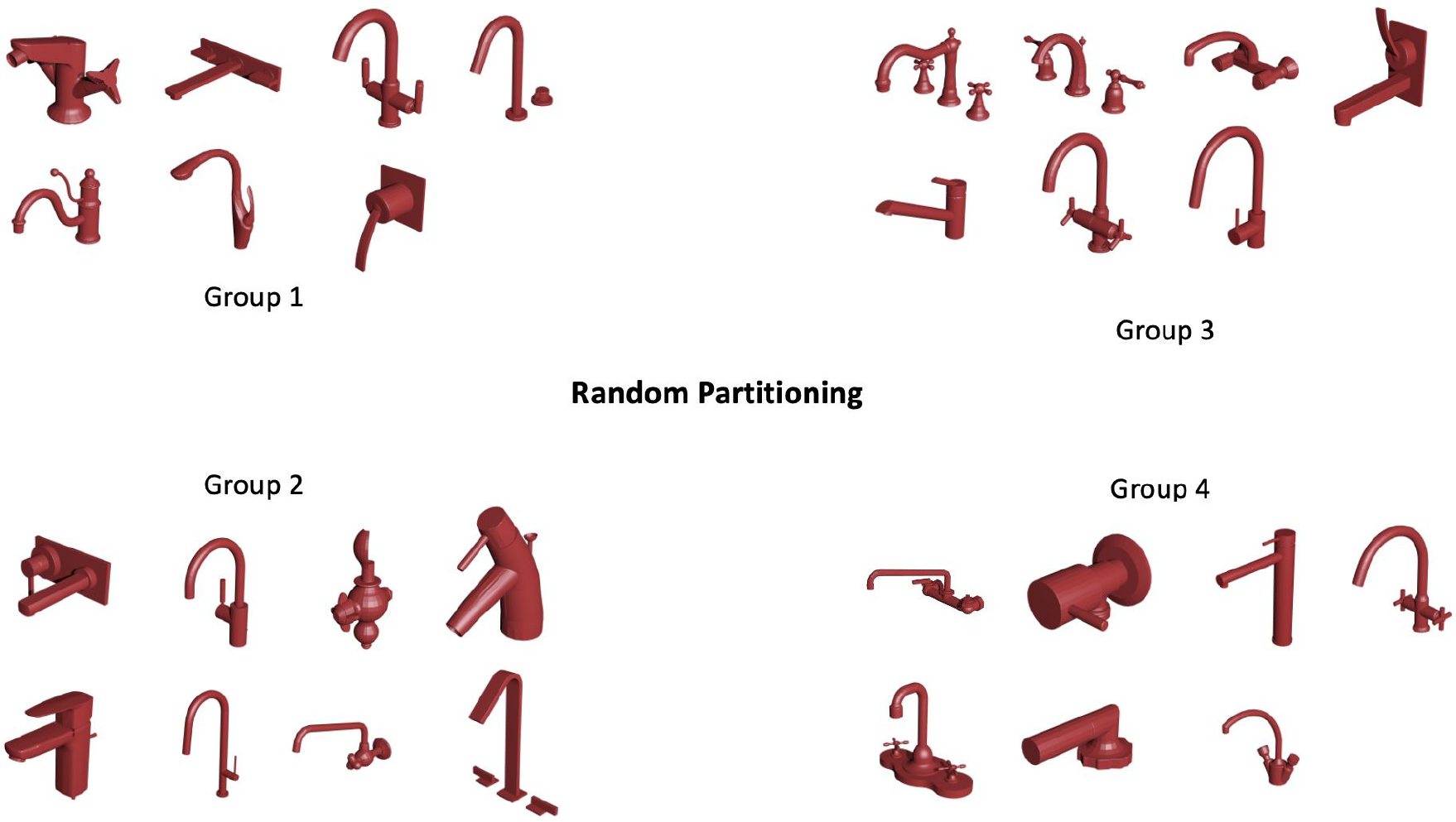}  % Replace with your image file
    \caption{An illustration of how random partitioning separates 29 low-performing faucets into 4 groups}
    \label{fig:random_partition}
\end{figure}

Fig \ref{fig:point_cloud_partition} illustrates the point cloud feature-based partitioning, organizing 29 low-performing faucets into four distinct groups. Faucets within the same cluster show clear similarities in design, though balanced clustering may slightly affect the precision of these groupings. For instance, faucets in Group 1 typically have two separate knobs positioned on either side of an arched spout. In contrast, Group 2 faucets feature a horizontally placed knob atop the spout. Group 3 faucets are often wall-mounted, with a vertically aligned knob, while Group 4 faucets are characterized by long, thin, curved spouts with the knob integrated into the spout itself. In contrast, Fig \ref{fig:random_partition} presents the result of random partitioning, where no shared characteristics are observed within each group.

With similar faucets grouped through point cloud feature-based partitioning, a specialist only needs to learn a single, consistent method of operating the faucets, significantly reducing the complexity of the learning process. For example, the specialist assigned to Group 1 in Fig \ref{fig:point_cloud_partition} only needs to focus on operating the knobs positioned on either side of the spout, while the specialist assigned to Group 2 only needs to operate the knob located atop the spout. In contrast, with the random partitioning shown in Fig \ref{fig:random_partition}, where faucets in each group vary, the specialist must learn all different methods of turning the faucets, which inevitably increases the difficulty of learning.

\section{Conclusions, Discussions, \& Limitations}

In this paper, we address the limitations of random task partitioning used in the vanilla Generalist-Specialist Learning (GSL) framework. We propose an enhancement to GSL by introducing point cloud feature-based task partitioning for robotic manipulation tasks involving object variations. Our approach first extracts point cloud features from object variations using a pre-trained PointNet++ model, normalizes them using Euclidean distance, and reduces dimensionality via PCA. It then applies balanced clustering with a greedy algorithm to group the point cloud feature vectors, assigning each cluster to a specialist. Evaluating our method on the challenging turn faucet task, which includes 60 different types of faucets from the ManiSkill benchmark, we demonstrate that our approach outperforms random partitioning by 9.4\% with a fixed number of specialists and reduces computation and sample requirements by 50\% to achieve similar performance.

\paragraph{\textbf{Limitations}} We focus on robotic manipulation tasks involving object variations. For tasks with other types of environmental variations, such as Procgen tasks, alternative heuristic task partitioning methods would need to be developed.

\clearpage
\bibliography{iclr2023_conference}
\bibliographystyle{iclr2023_conference}

\clearpage

\appendix
\textbf{Appendix}

\section{Key Hyperparameters}

The important hyperparameters used in PPO are shown in Table \ref{tab:hyperparameters}.

\begin{table}[h]
\centering
\begin{tabular}{lccl}
\toprule
\textbf{Hyperparameter} & \textbf{Values}\\
\midrule
Gamma & 0.99 \\
Update Epochs & 20 \\
Entropy Coefficient & 0.0 \\
Target KL & 0.1 \\
GAE Lambda & 0.9 \\
Learning Rate & 3e-4 \\
Minibatch Size & 400 \\
Num Steps Per Collect & 4000 \\
Value Loss Coefficient & 0.5 \\
Update-To-Data Ratio & 0.05 \\
\bottomrule
\end{tabular}
\caption{Key hyperparameters used in PPO algorithm}
\label{tab:hyperparameters}
\end{table}

\end{document}